# Sensorformer: Cross-patch attention with global-patch compression is effective for high-dimensional multivariate time series forecasting

| Liyang Qin | Xiaoli Wang | Chunhua Yang | Huaiwen Zou | Haochuan Zhang |
|---|---|---|---|---|
| School of Automation | School of Automation | School of Automation | School of Automation | School of Automation |
| Central South University | Central South University | Central South University | Central South University | Central South University |
| Changsha, China | Changsha, China | Changsha, China | Changsha, China | Changsha, China |
| qly520@csu.edu.cn | xlwang@csu.edu.cn | ychh@csu.edu.cn | 244603020@csu.edu.cn | 244607004@csu.edu.cn |

**Abstract**

Among the existing Transformer-based multivariate time series forecasting methods, iTransformer, which treats each variable sequence as a token and only explicitly extracts cross-variable dependencies, and PatchTST, which adopts a channel-independent strategy and only explicitly extracts cross-time dependencies, both significantly outperform most Channel-Dependent Transformer that simultaneously extract cross-time and cross-variable dependencies. This indicates that existing Transformer-based multivariate time series forecasting methods still struggle to effectively fuse these two types of information. We attribute this issue to the dynamic time lags in the causal relationships between different variables. Therefore, we propose a new multivariate time series forecasting Transformer, Sensorformer, which first compresses the global patch information and then simultaneously extracts cross-variable and cross-time dependencies from the compressed representations. Sensorformer can effectively capture the correct inter-variable correlations and causal relationships, even in the presence of dynamic causal lags between variables, while also reducing the computational complexity of pure cross-patch self-attention from $O(D^2 \cdot Patch\_num^2 \cdot d\_model)$ to $O(D^2 \cdot Patch\_num \cdot d\_model)$. Extensive comparative and ablation experiments on 9 mainstream real-world multivariate time series forecasting datasets demonstrate the superiority of Sensorformer. **The implementation of Sensorformer, following the style of the Time-series-library and scripts for reproducing the main results, is publicly available at https://github.com/BigYellowTiger/Sensorformer**

## 1. Introduction

Transformer[1], known for its ability to efficiently capture dependencies among global features and its excellent scalability, has achieved significant success in the fields of natural language processing[1][2][3] and computer vision[4][5][6]. However, in the early exploration of multivariate time series forecasting tasks, the Transformer did not demonstrate significant superiority[7]. Nevertheless, Nie et al.[8] soon revealed that a major reason for this issue lies in the single-point token construction method adopted by most approaches (as shown in Fig. 1(b1)), which struggles to represent key temporal features such as trends and distributions within a single token. To address this problem, PatchTST was proposed in [8], which significantly improved the performance of the Transformer in multivariate time series forecasting through the use of patch tokens and a channel-independent (CI) forward propagation strategy (as shown in Fig. 1(b3)). Subsequently, other patch-based multivariate time series forecasting Transformers, such as Crossformer[9] and TimeXer[10], have also achieved competitive performance.

However, according to common assumptions in previous studies, the explicit extraction of cross-variable dependencies is crucial for multivariate time series modeling, as there are often correlations or causal relationships between variables. Research [11] and [12] conducted a more detailed comparison between CI and channel-dependent strategies(CD), and the results revealed that, on almost all deep neural network backbones, including Transformers, the generalization ability of methods based on CI strategy significantly outperforms most CD-based methods. The conclusions of these studies suggest that the CI strategy should become the primary approach for multivariate time series forecasting. However, iTransformer[13], a method that treats each variable sequence as a token and only explicitly extracts cross-variable dependencies, has outperformed PatchTST on many mainstream datasets[10][13][14], achieving SOTA performance.

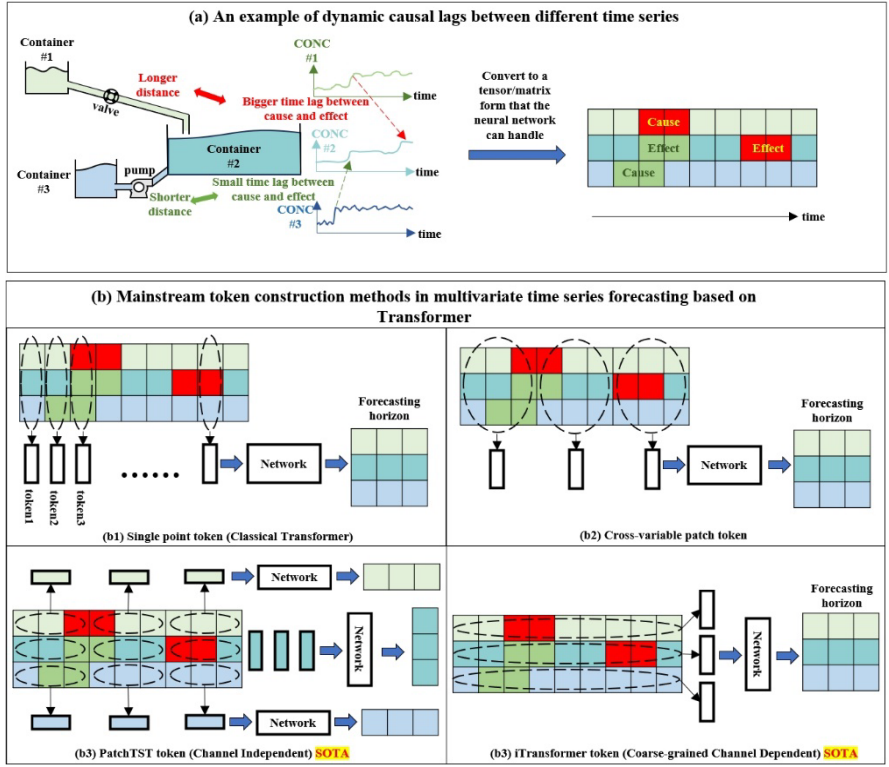

Fig. 1. Causal lag in multivariate time series and the mainstream token construction method in existing Transformer-based multivariate time series forecasting methods

Only extracting explicit cross-variable dependencies or only extracting cross-time dependencies can achieve good performance, indicating that both types of information are crucial for efficient multivariate time series forecasting. However, many CD Transformers, which attempt to balance cross-time and cross-variable dependencies (Fig.1(b1) and Fig.1(b2)), perform mediocrely[9][1]. This suggests that current Transformer-based methods for multivariate time series prediction cannot effectively fuse these two types of information. Zhou et al.[14] identified that forcibly fusing a large number of causally unrelated variables into a single token is a major cause of this issue. Besides this reason, based on our industrial research and experience, we believe that the **dynamic time lag of causal relationships between variables** is also a significant factor limiting the fusion of cross-variable and cross-time dependencies in existing Transformer-based methods.

Fig. 1 illustrates an example of how causal time lag between variables affects dependency extraction: the concentrations in containers #1 and #3 are causally related to the concentration in container #2. Due to differences in spatial distance, this causal relationship exhibits a dynamic time lag, reflected in the input tensor as misaligned data points between cause and effect. As shown in Fig. 1(b), classical CD Transformers (Figures 1(b1) and 1(b2)) are likely to forcibly fuse causally unrelated cross-variable data into a single token. In contrast, iTransformer and PatchTST can partially avoid the aforementioned issues, which may be a key factor in their state-of-the-art performance. However, both methods struggle to effectively balance the extraction of cross-time and cross-variable dependencies during the learning process.

The aforementioned analysis indicates that reducing the proportion of causally irrelevant data in each data fusion is key to improve the performance of Transformer-based multivariate time series forecasting methods. Therefore, we propose Sensorformer, which utilizes a two-stage pure cross-patch attention (Sensor Attention Block) to address the aforementioned issues. The main contributions of this paper are as follows:

(1) We analyzed one of the main limitations of existing Transformer-based multivariate time series forecasting methods, which is the forced fusion of causally unrelated cross-variable data due to dynamic time lags between variables.

(2) We propose a Transformer-based multivariate time series prediction method called Sensorformer, which utilizes a two-stage pure cross-patch attention mechanism. First, it compresses global patch information into a few representation vectors. Then, based on these compressed representations, it simultaneously extracts global cross-time and cross-variable dependencies. This innovative approach not only effectively reduces the computational complexity and CUDA memory usage of pure cross-patch attention when handling high-dimensional multivariate time series, but also significantly enhances the network's generalization ability.

(3) Experiments on nine mainstream multivariate time series datasets show that Sensorformer outperforms iTransformer and PatchTST, among other SOTA methods, in most cases. Additionally, its computational efficiency is significantly better than PatchTST and Transformer that uses pure cross-patch self-attention when processing high-dimensional datasets.

## 2. Related works

Multivariate time series are crucial in fields like industry, finance, and medicine. Recently, deep learning-based time series forecasting methods have gained attention due to their powerful nonlinear fitting and end-to-end modeling capabilities. These methods can be categorized into five types based on their backbone networks: RNN, CNN, MLP, GNN and Transformer.

**RNN-based methods** (e.g., DeepAR[15], LSTNet[16], DeepTrends[17]) capture temporal dependencies well but suffer from low parallelism and gradient vanishing in long sequences. **CNN-based methods** (e.g., TimesNet[18], SCINet[19], TemDep[20]) excel in multi-scale feature capture and local dependency analysis but require many layers, risking overfitting with small samples. **GNN-based methods** (e.g., TPGNN[21], Graph WaveNet[22]) are effective for cross-variable dependencies but need strong prior knowledge and have low computational efficiency. **MLP-based methods** (e.g., N-BEATS[23], DLinear[24]) are robust and efficient but lack support for variable-length sequences and need additional feature construction.

The **Transformer model**, initially for NLP, addresses information loss and parallel computation issues in long sequences through its self-attention mechanism[1]. Inspired by its success in NLP and CV, researchers tried apply it to time series forecasting tasks. The study[24] first used the pre-training and fine-tuning strategy with the Transformer in time series forecasting task, achieving competitive performance. Subsequent models like Informer[25], AutoFormer[26], and Pyraformer[27] effectively improved computational efficiency, while FedFormer[28] extended the frequency domain analysis capabilities of Transformer in time series forecasting task. Crossformer[9] highlighted the need for cross-variable dependency extraction, which further explored by TimeXer[10]. Despite initial skepticism[7], methods such as PatchTST[8] and iTransformer[13] have significantly improved the performance of Transformers in multivariate time series forecasting tasks. However, studies [14][29] indicate that existing Transformer-based multivariate time series prediction methods still face many limitations when dealing with dynamic time lags and cross-variable data fusion. The optimal application mode of the attention mechanism in multivariate time series prediction tasks still requires further research and improvement.

## 3. Proposed method

The objective of the multivariate time series forecasting task based on deep neural networks can be defined as follows.

For a multivariate time series system containing $D$ observable variables, given the historical observations $X^{his} = \begin{bmatrix} x_1^{(1)} & x_1^{(2)} & \ldots & x_1^{(D)} \\ \ldots & \ldots & \ldots & \ldots \\ x_L^{(1)} & x_L^{(2)} & \ldots & x_L^{(D)} \end{bmatrix}$ of $L$ consecutive sampling points, where $x_i^{(D)} \in R$ represents the observation value of the D-th input variable at the i-th time point, a forecasting model $f(\cdot)$ with model parameters $\theta$ obtained through training can be used to forecast the values $X^{future} = \begin{bmatrix} x_{L+1}^{(1)} & x_{L+1}^{(2)} & \ldots & x_{L+1}^{(D)} \\ \ldots & \ldots & \ldots & \ldots \\ x_{L+H}^{(1)} & x_{L+H}^{(2)} & \ldots & x_{L+H}^{(D)} \end{bmatrix}$ of the system for the next $H$ time steps. This can be expressed as $X^{future} = f(X^{his}; \theta)$.

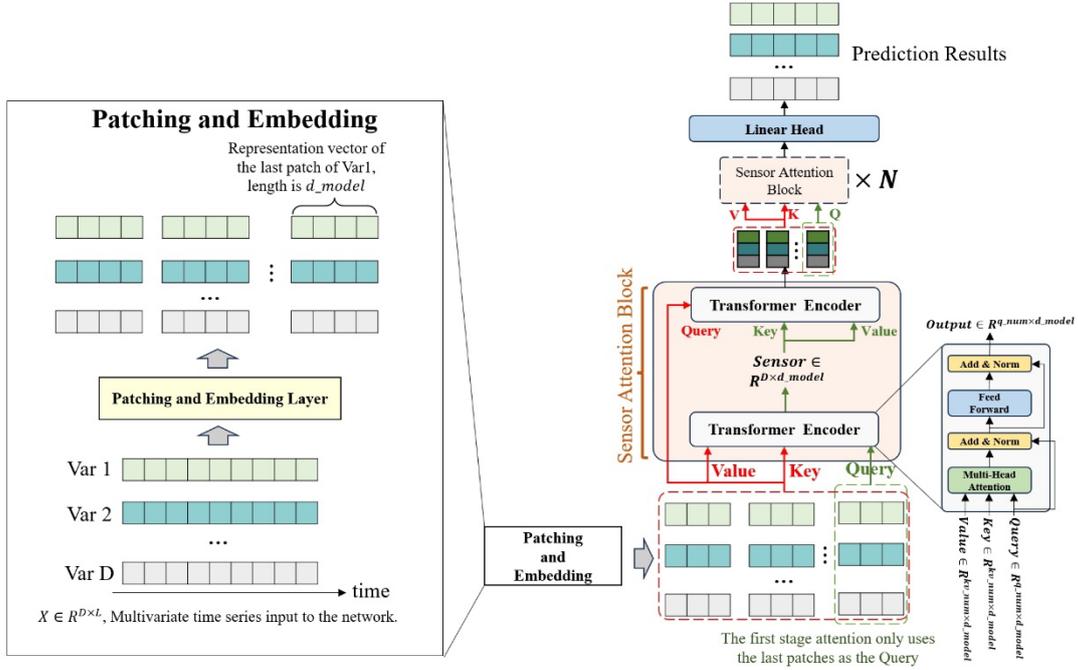

Fig. 2. Overall structure of Sensorformer

### 3.1 Overview of the Sensorformer

The overall structure of the proposed Sensorformer is shown in Fig. 2. The core idea of Sensorformer is very straightforward: since the characteristic of time series is to use historical information to predict the future, the last patch of each variable in the input sequence can be regarded as the result of accumulating all history causal relationships. Therefore, using the **last patch** of each variable as the Query and the all patches as the Key and Value for attention calculation, we can theoretically obtain **a compressed representation of historical causal relationships that is closest to the future value to be predicted in the time dimension.** We call this compressed representation $Sensor$. Then, by using the $Sensor$ as a knowledge base (i.e., Key and Value) and all patches as the Query for attention, the network can simultaneously capture cross-time dependencies and cross-variable dependencies from two directions while reducing computational complexity. The detailed description of the aforementioned process is provided in Sections 3.2 and 3.3.

### 3.2 Patching and embedding

Sensorformer adopts the same patch strategy as PatchTST[8]: suppose there are $D$ variable sequences, $X_{1:L}^{(i)} \in R^{L \times 1}$ denotes the i-th variable sequence, the length of the input lookback window is $L$, the length of each patch is $P$ and the stride is $S$. Then for each $X_{1:L}^{(i)}$, we first pad $S$ repeated number of the last value of $X_{1:L}^{(i)}$ to the end of $X_{1:L}^{(i)}$. Next, extract $N = \left\lfloor \frac{(L-P)}{S} \right\rfloor + 2$ subsequences from $X_{1:L}^{(i)}$, where the elements contained in the j-th subsequence are $[x_{j \times S}^{(i)}, x_{j \times S+1}^{(i)}, \ldots, x_{j \times S+P-1}^{(i)}]$. Each subsequence is called a patch, and each patch will be fed into the embedding layer, mapped to a representation vector of length $d\_model$, and augmented with positional encoding[1]. After patching and embedding, the original input $X_{1:L}^{(i)} \in R^{L \times 1}$ is mapped to $EPatches \in R^{D \times N \times d\_model}$, which serves as the input to the first Sensor Attention Block.

### 3.3 Sensor Attention Block

To avoid the forced fusion of causally unrelated cross-variable data due to dynamic causal lags between variables, it is intuitive to consider **treating all patches as queries, keys, and values for self-attention computation, that is, pure cross-patch self-attention**. This approach allows for the comparison of the similarity between any two time periods of any two variables. Regardless of the causal time lag between two variables, this method effectively merges the two patches with the strongest similarities. However, the time complexity within a single attention head for **pure cross-patch self-attention** is $O(D^2 \cdot N^2 \cdot d\_model)$, where $N$ is the number of patches. Therefore, we propose a two-stage attention block, named the Sensor Attention Block, which first compresses the representation of all patches and then extracts cross-variable and cross-time dependencies simultaneously. This method maintains fine-grained cross-patch attention while achieving a time complexity of only $O(D^2 \cdot N \cdot d\_model)$. The detail calculation process of the Sensor Attention Block is as follows:

**First-stage attention for compressing the representation of all input patches**: In the first-stage of attention, the last patches of all variables are treated as the Query, while all patches are treated as the Key and Value. A vanilla multi-head attention calculation[1] is then performed, and the output is referred to as the $Sensor$, which represents the compressed representation of the global patches. To facilitate understanding and avoid ambiguity, we describe the detailed steps of the first-stage attention through the pseudocode in Algorithm 1. The time complexity of the first-stage attention within a single attention head is $O(D^2 \cdot N \cdot d\_model)$.

| Algorithm 1. First-stage attention of Sensor Attention Block |
|---|
| **Input**: $EPatches \in R^{D \times N \times d\_model}$, the representation vectors of all patches, where $EPatches[i,j,:]$ represents the representation vector of the j-th patch in the i-th variable. |
| **Output**: $Sensor \in R^{D \times d\_model}$, $D$ vectors, the compressed representation of $EPatches$. |
| **Step 1.** $Query = EPatches[:,-1,:].view(D, d\_model)$, $Key = Value = EPatches.view(D \times N, d\_model)$ |
| **Step 2.** $Z^{stage1} = LayerNorm(Query + MHA(Query, Key, Value))$ |
| **Step 3.** $Sensor = LayerNorm(Z^{stage1} + MLP(Z^{stage1}))$ |
| **Step 3.** Output $Sensor$ |

$LayerNorm(\cdot)$ refers to the layer normalization module, proposed by [31], which is utilized to

smooth feature distributions and enhance nonlinearity. $MLP(\cdot)$ denotes a multi-layer perceptron network. $MHA(\cdot)$ represents the multi-head attention module[1]. The computation process within each attention head of $MHA(Query, Key, Value)$ is as follows: Given the $Query \in R^{Q\_num \times d\_model}$ and $Key, Value \in R^{KV\_num \times d\_model}$ respectively, the attention weights are first computed as $W = Softmax\left(\frac{Query \cdot Key^T}{\sqrt{d_{model}}}\right)$, where $W \in R^{Q\_num \times KV\_num}$. The j-th element in the i-th row vector of $W$ represents the correlation between the i-th row vector in $Query$ (a token's representation vector) and j-th row vector in $Key$ and $Value$. Therefore, for the i-th row vector $Query[i,:]$ in $Query$, the representation of $Query[i,:]$ using $Value$ is $Rep(Query[i,:]) = \sum_{j=1}^{KV\_num} W[i,j] \cdot Vaule[j,:]$.

**Second-stage attention for extracting cross-variable and cross-time dependencies based on Sensor**: The second-stage attention mechanism treats $EPatches$ as the $Query$ and $Sensor$ as the $Key$ and $Value$. It utilizes cross-patch attention to simultaneously extract cross-variable and cross-time dependencies based on the compressed representation of $Sensor$. The detailed process is illustrated in the pseudocode of Algorithm 2. The time complexity of the second stage attention within a single attention head is $O(D^2 \cdot N \cdot d\_model)$.

| Algorithm 2. Second-stage attention of Sensor Attention Block |
|---|
| **Input**: $EPatches \in R^{D \times N \times d\_model}$, the representation vectors of all patches. |
| **Input**: $Sensor \in R^{D \times d\_model}$, $D$ vectors from the first-stage attention, the compressed representation of $EPatches$. |
| **Output**: $next\_EPatches \in R^{D \times N \times d\_model}$, the new representation vectors of all patches, the output of the current Sensor Attention Block, are used as input for the next layer of the Sensor Attention Block or the final prediction layer. |
| **Step 1.** $Query = EPatches.view(D \times N, d_{model})$, $Key = Value = Sensor$ |
| **Step 2.** $Z^{stage2} = LayerNorm(Query + MHA(Query, Key, Value))$ |
| **Step 3.** $next\_EPatches = LayerNorm(Z^{stage2} + MLP(Z^{stage2}))$ |
| **Step 3.** Output $next\_EPatches$ |

### 3.4 Output layer of Sensorformer

The output layer of Sensorformer is a linear layer with an input dimension of $N \times d\_model$ and an output dimension of $H$. The output of the final Sensor Attention Block is unfolded into D representation vectors, each of length $N \times d\_model$. Each representation vector is then fed into the output layer. The vector of length $H$ obtained from the output layer for the i-th representation vector is the prediction of Sensorformer for the i-th variable at the future $H$ time steps.

### 4. Experiments

The experimental section mainly includes the following contents: (1) Performance comparison of Sensorformer with state-of-the-art multivariate time series forecasting methods on 9 mainstream benchmark datasets; (2) Distribution of casual time lags between variables across different datasets; (3) Comparison of computational efficiency and CUDA memory usage between Sensorformer and SOTA Transformer methods; (4) Ablation studies, focusing on the impact of removing the first-stage attention and the second-stage attention in Sensorformer.

Details of all datasets and baseline methods used for comparison are provided in Appendix A. Experiments on the distribution of casual time lag across all datasets, the impact of major hyperparameters, the performance robustness of Sensorformer, and visualization of prediction results are provided in Appendices B.1, B.2, B.3 and B.4, respectively.

### 4.1 Performance comparison of Sensorformer with state-of-the-art multivariate time series

**forecasting methods**

Table 1 shows the MSE (Mean Squared Error), MAE (Mean Absolute Error), and the number of times the lowest and second lowest errors were achieved for all models. **The results shows that Sensorformer achieved the lowest and second lowest errors the most times, with a total of 64 times.** iTransformer and PatchTST ranked second and third, respectively. This experimental result first demonstrates the significant superiority of Sensorformer in many types of multivariate time series forecasting tasks. Moreover, although both Sensorformer and Crossformer incorporate cross-variable dependency extraction in the Patch strategy, Sensorformer, by using pure cross-patch attention, avoids forcibly fusing causally unrelated cross-variable data. This also reveals a key area for further exploration in multivariate time series forecasting research based on Transformers, that is, how to improve the accuracy of causal relationship identification during cross-variable data fusion.

**Table 1.** Full prediction results, where Avg represents the average error of the four prediction lengths, red represents the lowest error for a specific dataset and prediction horizon, blue represents the second lowest error, **1st Count** represents the number of times the corresponding method achieved the lowest error, and **2nd Count** represents the number of times it achieved the second lowest error.

| Models | | Sensorformer (Ours) | | iTransformer | | PatchTST | | Crossformer | | TiDE | | TimesNet | | Dlinear | | SCINet | | FEDformer | | Stationary | | Autoformer | |
|---|---|---|---|---|---|---|---|---|---|---|---|---|---|---|---|---|---|---|---|---|---|---|---|
| Metric | | MSE | MAE | MSE | MAE | MSE | MAE | MSE | MAE | MSE | MAE | MSE | MAE | MSE | MAE | MSE | MAE | MSE | MAE | MSE | MAE | MSE | MAE |
| ETTm1 | 96 | 0.329 | 0.369 | 0.334 | 0.368 | 0.334 | 0.374 | 0.404 | 0.426 | 0.364 | 0.387 | 0.338 | 0.375 | 0.345 | 0.372 | 0.418 | 0.438 | 0.379 | 0.419 | 0.386 | 0.398 | 0.505 | 0.475 |
| | 192 | 0.371 | 0.389 | 0.374 | 0.389 | 0.376 | 0.393 | 0.45 | 0.451 | 0.398 | 0.404 | 0.374 | 0.387 | 0.38 | 0.389 | 0.439 | 0.45 | 0.426 | 0.441 | 0.459 | 0.444 | 0.553 | 0.496 |
| | 336 | 0.4 | 0.41 | 0.426 | 0.42 | 0.404 | 0.413 | 0.532 | 0.515 | 0.428 | 0.425 | 0.41 | 0.411 | 0.413 | 0.413 | 0.49 | 0.485 | 0.445 | 0.459 | 0.495 | 0.464 | 0.621 | 0.537 |
| | 720 | 0.459 | 0.442 | 0.491 | 0.459 | 0.468 | 0.447 | 0.666 | 0.589 | 0.487 | 0.461 | 0.478 | 0.45 | 0.474 | 0.453 | 0.595 | 0.55 | 0.543 | 0.49 | 0.585 | 0.516 | 0.671 | 0.561 |
| | Avg | 0.39 | 0.403 | 0.406 | 0.409 | 0.396 | 0.407 | 0.513 | 0.496 | 0.419 | 0.419 | 0.4 | 0.406 | 0.403 | 0.407 | 0.485 | 0.481 | 0.448 | 0.452 | 0.481 | 0.456 | 0.588 | 0.517 |
| ETTm2 | 96 | 0.176 | 0.263 | 0.18 | 0.263 | 0.176 | 0.262 | 0.287 | 0.366 | 0.207 | 0.305 | 0.187 | 0.267 | 0.193 | 0.292 | 0.286 | 0.377 | 0.203 | 0.287 | 0.192 | 0.274 | 0.255 | 0.339 |
| | 192 | 0.246 | 0.306 | 0.25 | 0.309 | 0.243 | 0.306 | 0.414 | 0.492 | 0.29 | 0.364 | 0.249 | 0.309 | 0.284 | 0.362 | 0.399 | 0.445 | 0.269 | 0.328 | 0.28 | 0.339 | 0.281 | 0.34 |
| | 336 | 0.311 | 0.351 | 0.311 | 0.348 | 0.306 | 0.346 | 0.597 | 0.542 | 0.377 | 0.422 | 0.321 | 0.351 | 0.369 | 0.427 | 0.637 | 0.591 | 0.325 | 0.366 | 0.334 | 0.361 | 0.339 | 0.372 |
| | 720 | 0.413 | 0.408 | 0.412 | 0.406 | 0.407 | 0.405 | 1.73 | 1.042 | 0.558 | 0.524 | 0.408 | 0.403 | 0.554 | 0.522 | 0.96 | 0.735 | 0.421 | 0.415 | 0.417 | 0.413 | 0.433 | 0.432 |
| | Avg | 0.286 | 0.332 | 0.288 | 0.332 | 0.283 | 0.33 | 0.757 | 0.61 | 0.358 | 0.404 | 0.291 | 0.333 | 0.35 | 0.401 | 0.571 | 0.537 | 0.305 | 0.349 | 0.306 | 0.347 | 0.327 | 0.371 |
| ETTh1 | 96 | 0.381 | 0.4 | 0.386 | 0.405 | 0.414 | 0.419 | 0.423 | 0.448 | 0.479 | 0.464 | 0.384 | 0.402 | 0.386 | 0.4 | 0.654 | 0.599 | 0.376 | 0.419 | 0.513 | 0.491 | 0.449 | 0.459 |
| | 192 | 0.43 | 0.428 | 0.441 | 0.436 | 0.46 | 0.445 | 0.471 | 0.474 | 0.525 | 0.492 | 0.436 | 0.429 | 0.437 | 0.432 | 0.719 | 0.631 | 0.42 | 0.448 | 0.534 | 0.504 | 0.5 | 0.482 |
| | 336 | 0.465 | 0.446 | 0.487 | 0.458 | 0.501 | 0.466 | 0.57 | 0.546 | 0.565 | 0.515 | 0.491 | 0.469 | 0.481 | 0.459 | 0.778 | 0.659 | 0.459 | 0.465 | 0.588 | 0.535 | 0.521 | 0.496 |
| | 720 | 0.504 | 0.488 | 0.503 | 0.491 | 0.5 | 0.488 | 0.653 | 0.621 | 0.594 | 0.558 | 0.521 | 0.5 | 0.519 | 0.516 | 0.836 | 0.699 | 0.506 | 0.507 | 0.643 | 0.616 | 0.514 | 0.512 |
| | Avg | 0.445 | 0.441 | 0.454 | 0.447 | 0.469 | 0.454 | 0.529 | 0.522 | 0.541 | 0.507 | 0.458 | 0.45 | 0.456 | 0.452 | 0.747 | 0.647 | 0.44 | 0.46 | 0.57 | 0.537 | 0.496 | 0.487 |
| ETTh2 | 96 | 0.291 | 0.343 | 0.297 | 0.348 | 0.302 | 0.348 | 0.745 | 0.584 | 0.4 | 0.44 | 0.34 | 0.374 | 0.333 | 0.387 | 0.707 | 0.621 | 0.358 | 0.397 | 0.476 | 0.458 | 0.346 | 0.388 |
| | 192 | 0.373 | 0.394 | 0.38 | 0.4 | 0.388 | 0.4 | 0.877 | 0.656 | 0.528 | 0.509 | 0.402 | 0.414 | 0.477 | 0.476 | 0.86 | 0.689 | 0.429 | 0.439 | 0.512 | 0.493 | 0.456 | 0.452 |
| | 336 | 0.426 | 0.43 | 0.428 | 0.432 | 0.426 | 0.433 | 1.043 | 0.731 | 0.643 | 0.571 | 0.452 | 0.452 | 0.594 | 0.541 | 1 | 0.744 | 0.496 | 0.487 | 0.552 | 0.551 | 0.482 | 0.486 |
| | 720 | 0.432 | 0.448 | 0.431 | 0.448 | 0.432 | 0.446 | 1.104 | 0.763 | 0.874 | 0.679 | 0.462 | 0.468 | 0.831 | 0.657 | 1.249 | 0.838 | 0.463 | 0.474 | 0.562 | 0.56 | 0.515 | 0.511 |
| | Avg | 0.381 | 0.404 | 0.384 | 0.407 | 0.387 | 0.407 | 0.942 | 0.684 | 0.611 | 0.55 | 0.414 | 0.427 | 0.559 | 0.515 | 0.954 | 0.723 | 0.437 | 0.449 | 0.526 | 0.516 | 0.45 | 0.459 |
| ECL | 96 | 0.19 | 0.279 | 0.187 | 0.268 | 0.193 | 0.281 | 0.219 | 0.314 | 0.237 | 0.329 | 0.168 | 0.272 | 0.199 | 0.282 | 0.247 | 0.345 | 0.193 | 0.308 | 0.169 | 0.273 | 0.201 | 0.317 |
| | 192 | 0.195 | 0.285 | 0.194 | 0.278 | 0.196 | 0.285 | 0.231 | 0.322 | 0.236 | 0.33 | 0.184 | 0.289 | 0.196 | 0.285 | 0.257 | 0.355 | 0.201 | 0.315 | 0.182 | 0.286 | 0.222 | 0.334 |
| | 336 | 0.212 | 0.302 | 0.212 | 0.296 | 0.212 | 0.301 | 0.246 | 0.337 | 0.249 | 0.344 | 0.198 | 0.3 | 0.209 | 0.301 | 0.269 | 0.369 | 0.214 | 0.329 | 0.2 | 0.304 | 0.231 | 0.338 |
| | 720 | 0.254 | 0.334 | 0.256 | 0.332 | 0.254 | 0.334 | 0.28 | 0.363 | 0.284 | 0.373 | 0.22 | 0.32 | 0.245 | 0.333 | 0.299 | 0.39 | 0.246 | 0.355 | 0.222 | 0.321 | 0.254 | 0.361 |
| | Avg | 0.213 | 0.3 | 0.212 | 0.294 | 0.214 | 0.3 | 0.244 | 0.334 | 0.251 | 0.344 | 0.192 | 0.295 | 0.212 | 0.3 | 0.268 | 0.365 | 0.214 | 0.327 | 0.193 | 0.296 | 0.227 | 0.338 |
| Exchange | 96 | 0.086 | 0.204 | 0.086 | 0.205 | 0.087 | 0.206 | 0.256 | 0.367 | 0.094 | 0.218 | 0.107 | 0.234 | 0.088 | 0.218 | 0.267 | 0.396 | 0.148 | 0.278 | 0.111 | 0.237 | 0.197 | 0.323 |
| | 192 | 0.184 | 0.302 | 0.176 | 0.302 | 0.179 | 0.301 | 0.47 | 0.509 | 0.184 | 0.307 | 0.226 | 0.344 | 0.176 | 0.315 | 0.351 | 0.459 | 0.271 | 0.315 | 0.219 | 0.335 | 0.3 | 0.369 |
| | 336 | 0.334 | 0.418 | 0.336 | 0.422 | 0.339 | 0.422 | 1.268 | 0.883 | 0.349 | 0.431 | 0.367 | 0.448 | 0.313 | 0.427 | 1.324 | 0.853 | 0.46 | 0.427 | 0.421 | 0.476 | 0.509 | 0.524 |
| | 720 | 0.851 | 0.697 | 0.847 | 0.693 | 0.855 | 0.698 | 1.767 | 1.068 | 0.852 | 0.698 | 0.964 | 0.746 | 0.839 | 0.695 | 1.058 | 0.797 | 1.195 | 0.695 | 1.092 | 0.769 | 1.447 | 0.941 |
| | Avg | 0.364 | 0.406 | 0.361 | 0.406 | 0.365 | 0.406 | 0.94 | 0.707 | 0.37 | 0.413 | 0.416 | 0.443 | 0.354 | 0.414 | 0.75 | 0.626 | 0.519 | 0.429 | 0.461 | 0.454 | 0.613 | 0.539 |
| Traffic | 96 | 0.507 | 0.339 | 0.511 | 0.345 | 0.548 | 0.353 | 0.811 | 0.503 | 0.805 | 0.493 | 0.593 | 0.321 | 0.65 | 0.396 | 0.788 | 0.499 | 0.587 | 0.366 | 0.612 | 0.338 | 0.613 | 0.388 |
| | 192 | 0.506 | 0.338 | 0.519 | 0.348 | 0.534 | 0.345 | 0.817 | 0.505 | 0.756 | 0.474 | 0.617 | 0.336 | 0.598 | 0.37 | 0.789 | 0.505 | 0.604 | 0.373 | 0.613 | 0.34 | 0.616 | 0.382 |
| | 336 | 0.521 | 0.343 | 0.539 | 0.359 | 0.546 | 0.349 | 0.83 | 0.517 | 0.762 | 0.477 | 0.629 | 0.336 | 0.605 | 0.373 | 0.797 | 0.508 | 0.621 | 0.383 | 0.618 | 0.328 | 0.622 | 0.337 |
| | 720 | 0.555 | 0.349 | 0.565 | 0.378 | 0.582 | 0.366 | 0.839 | 0.528 | 0.719 | 0.449 | 0.64 | 0.35 | 0.645 | 0.394 | 0.841 | 0.523 | 0.626 | 0.382 | 0.653 | 0.355 | 0.66 | 0.408 |
| | Avg | 0.523 | 0.345 | 0.534 | 0.358 | 0.553 | 0.353 | 0.824 | 0.513 | 0.76 | 0.473 | 0.62 | 0.336 | 0.625 | 0.383 | 0.804 | 0.509 | 0.61 | 0.376 | 0.624 | 0.34 | 0.628 | 0.379 |
| Weather | 96 | 0.186 | 0.223 | 0.189 | 0.229 | 0.184 | 0.224 | 0.19 | 0.266 | 0.202 | 0.261 | 0.172 | 0.22 | 0.196 | 0.255 | 0.221 | 0.306 | 0.217 | 0.296 | 0.173 | 0.223 | 0.266 | 0.336 |
| | 192 | 0.232 | 0.263 | 0.235 | 0.266 | 0.232 | 0.263 | 0.236 | 0.266 | 0.242 | 0.298 | 0.219 | 0.261 | 0.237 | 0.296 | 0.261 | 0.34 | 0.276 | 0.336 | 0.245 | 0.285 | 0.307 | 0.367 |
| | 336 | 0.285 | 0.302 | 0.288 | 0.306 | 0.285 | 0.303 | 0.292 | 0.323 | 0.287 | 0.335 | 0.28 | 0.306 | 0.283 | 0.335 | 0.309 | 0.378 | 0.339 | 0.38 | 0.321 | 0.338 | 0.359 | 0.395 |
| | 720 | 0.358 | 0.35 | 0.369 | 0.353 | 0.361 | 0.351 | 0.39 | 0.412 | 0.351 | 0.386 | 0.365 | 0.359 | 0.345 | 0.381 | 0.377 | 0.427 | 0.403 | 0.428 | 0.414 | 0.41 | 0.419 | 0.428 |
| | Avg | 0.265 | 0.285 | 0.27 | 0.289 | 0.266 | 0.285 | 0.277 | 0.312 | 0.271 | 0.32 | 0.259 | 0.287 | 0.265 | 0.317 | 0.292 | 0.363 | 0.309 | 0.36 | 0.288 | 0.314 | 0.338 | 0.382 |
| Solar-Energy | 96 | 0.233 | 0.276 | 0.235 | 0.276 | 0.239 | 0.281 | 0.31 | 0.331 | 0.312 | 0.399 | 0.25 | 0.292 | 0.29 | 0.378 | 0.237 | 0.344 | 0.242 | 0.342 | 0.215 | 0.249 | 0.884 | 0.711 |
| | 192 | 0.266 | 0.297 | 0.279 | 0.304 | 0.276 | 0.305 | 0.734 | 0.725 | 0.339 | 0.416 | 0.296 | 0.318 | 0.32 | 0.398 | 0.28 | 0.38 | 0.285 | 0.38 | 0.254 | 0.272 | 0.834 | 0.692 |
| | 336 | 0.292 | 0.312 | 0.311 | 0.325 | 0.303 | 0.321 | 0.75 | 0.735 | 0.368 | 0.43 | 0.319 | 0.33 | 0.353 | 0.415 | 0.304 | 0.389 | 0.282 | 0.376 | 0.29 | 0.296 | 0.941 | 0.723 |
| | 720 | 0.292 | 0.313 | 0.311 | 0.325 | 0.304 | 0.319 | 0.769 | 0.765 | 0.37 | 0.425 | 0.338 | 0.337 | 0.356 | 0.413 | 0.308 | 0.388 | 0.357 | 0.427 | 0.285 | 0.295 | 0.882 | 0.717 |
| | Avg | 0.271 | 0.3 | 0.284 | 0.308 | 0.281 | 0.307 | 0.641 | 0.639 | 0.347 | 0.417 | 0.301 | 0.319 | 0.33 | 0.401 | 0.282 | 0.375 | 0.291 | 0.381 | 0.261 | 0.278 | 0.885 | 0.711 |
| 1st Count | | 16 | 19 | 3 | 7 | 7 | 2 | 0 | 0 | 0 | 0 | 8 | 8 | 0 | 0 | 0 | 0 | 4 | 0 | 5 | 5 | 0 | 0 |
| 2nd Count | | 15 | 14 | 16 | 17 | 8 | 11 | 0 | 0 | 1 | 1 | 3 | 11 | 2 | 3 | 0 | 0 | 0 | 1 | 5 | 3 | 0 | 0 |
| Sum of 1st and 2nd | | 64 | | 43 | | 35 | | 0 | | 2 | | 30 | | 11 | | 0 | | 5 | | 18 | | 0 | |

### 4.2 Time lag in similarity between variables in different datasets

The introduction analyzes that the dynamic time lag between variables is one of the main reasons limiting many Transformer-based multivariate time series forecasting methods. **To demonstrate that the dynamic time lag in variable correlation is a common phenomenon**, we visually present the proportion of variable pairs with casual time lag in the Weather, ECL, and Traffic datasets. The comparison method is as follows: (1) randomly select an input tensor; (2) after performing patching operations on the input tensor, we take the middle patch of the i-th variable (the 5-th patch due to consistent dataset division) in this tensor as the base patch, then calculate the Pearson Correlation Coefficient (PCC) between the base patch and all patches of the j-th variable; (3) if the patch with the maximum PCC in the j-th variable sequence has a different time index from the base patch, it is considered that there is a certain degree of time lag in the correlation or causality between variable i and variable j; (4) finally, calculate the proportion of variable pairs with causal time lag to the total number of variable pairs in the current input tensor. The experimental results are shown in Table 2 and Fig. 3. Complete experimental results on the time lag between variables in different datasets are provided in Appendix B.1. The experimental results in Table 2 indicate that in the three datasets, the majority of variable pairs (averaging over 50% across the three datasets) exhibit a certain degree of time lag in their maximum PCC. Although the results in Table 2, Fig.3 and Appendix B.1 do not fully prove the existence of time lag in the causal relationship, it strongly indicates the necessity of incorporating time lag analysis capabilities in the network when conducting cross-variable similarity analysis.

**Table 2. Proportion of variable pairs with casual time lag**

| Datasets | Weather | ECL | Traffic |
| --- | --- | --- | --- |
| Input tensor 1 | 54.29% | 72.34% | 75.71% |
| Input tensor 2 | 71.42% | 81.92% | 61.18% |
| Input tensor 3 | 73.33% | 63.88% | 16.10% |
| Input tensor 4 | 85.71% | 70.6% | 75.24% |
| Input tensor 5 | 63.34% | 71.05% | 52.32% |
| Avg | 69.62% | 71.96% | 56.11% |

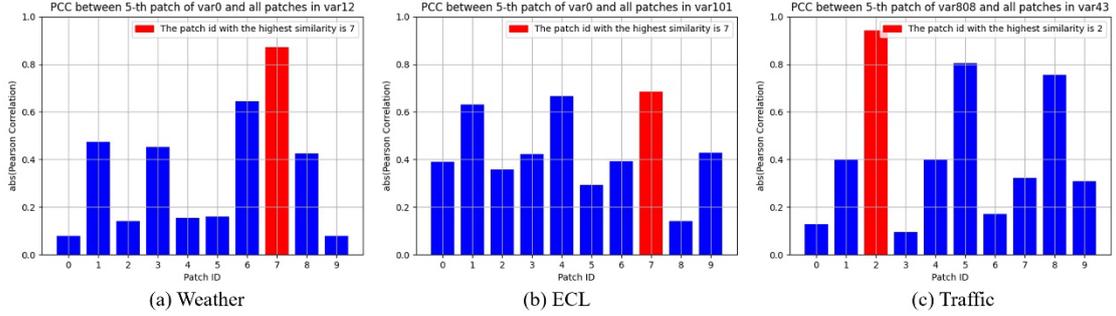

Fig. 3 Example of PCC distribution between two random variables within an input tensor in the Weather, ECL, and Traffic datasets

### 4.3 Comparison of computational efficiency and CUDA memory usage

Section 3.3 mentions that using pure cross-patch self-attention, where all patches are treated as queries, keys, and values, can also handle the situation of dynamic causal time lags between variables. However, its computational complexity reaches $O(D^2 \cdot N^2 \cdot d\_model)$. In contrast, Sensorformer reduces the complexity to $O(D^2 \cdot N \cdot d\_model)$ by first compressing the global patches and then extracting dependencies. To visually demonstrate the efficiency advantages of Sensorformer's two-stage attention, we compared the training time and CUDA memory usage of Sensorformer, pure cross-patch self-attention, iTransformer, PatchTST and Crossformer on three high-dimensional datasets: Solar-Energy, ECL, and Traffic. The experimental results are shown in Fig. 4.

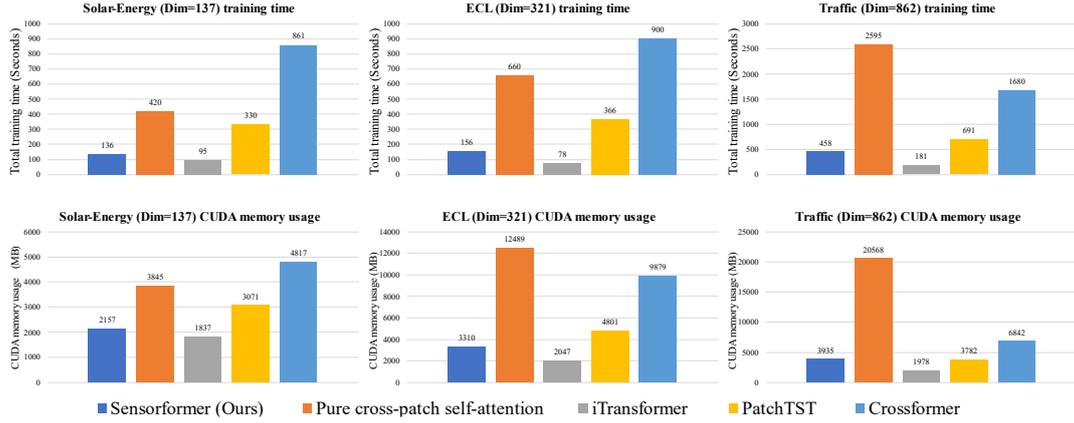

Fig. 4 Comparison of computational efficiency and CUDA memory usage of Sensorformer with pure-cross patch self-attention, iTransformer, PatchTST, and Crossformer on three high-dimensional datasets

Fig. 4 shows that the training time of Sensorformer is significantly lower than the other three methods except for iTransformer. Moreover, the CUDA memory usage of Sensorformer is only slightly higher than PatchTST in the Traffic dataset, while in other cases, it is significantly lower than the other three methods except for iTransformer. Additionally, it can be observed that as the dimensionality increases, the training time and memory usage of pure cross-patch self-attention show a sharp increase. This further demonstrates the effectiveness and necessity of introducing global information compression representations when dealing with high-dimensional multivariate time series.

**4.4 Ablation Study**

Since the main innovation of Sensorformer lies in the two-stage attention mechanism within the Sensor Attention Block, the ablation experiments mainly compare the impact of removing the first and second stages of attentiossn in the Sensor Attention Block on the ETTh1(low-dimensional dataset), Weather(medium-dimensional dataset) and ECL(high-dimensional dataset) datasets. Removing the first stage of Sensor Attention Block is equivalent to not compressing the global patches and directly performing **cross-patch self-attention**. Removing the second stage of attention means directly using the compressed representation of the global patches, $Sensor$, as the attention calculation result. However, directly removing the second stage of attention will lead to inconsistent input and output shapes of the Sensor Attention Block, making it impossible to stack directly. Therefore, after removing the second stage of attention, we replicate each representation vector in the Sensor $N$ times to ensure that the input and output tensor shapes of the attention calculation are consistent. The experimental results are shown in Fig. 5.

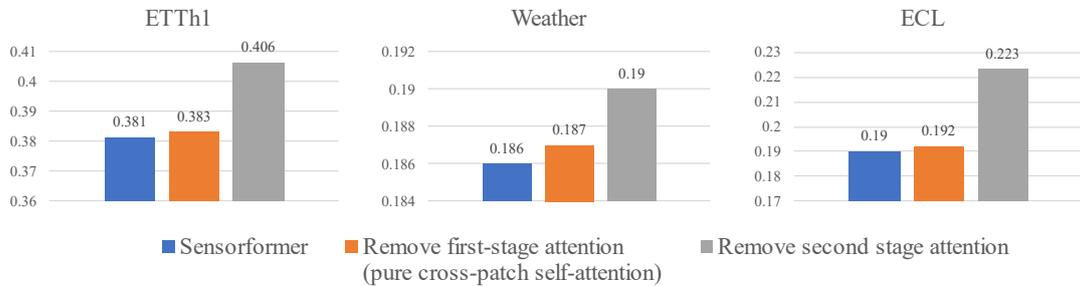

Fig. 5 Ablation experiment results

From Fig. 5, we can observe the following phenomena: (1) The increase in MSE when removing the second stage attention of Sensor Attention Block is significantly higher than when removing the first stage attention. This indicates that relying solely on the sensor (the representation of last patch of each variable) may harm learning ability. (2) After removing the first stage of attention, the MSE increases on all datasets. This shows that the compressed representation of Sensorformer helps the network more effectively extract dependencies between patches. Combining the significant advantages of Sensorformer in training time and CUDA memory usage compared to pure cross-patch self-attention (shown in Fig. 4), it can be fully demonstrated that compressing all patches into low-dimensional representations before dependency extraction is far superior to directly performing self-attention on all patches. This phenomenon also reveals an interesting point: **Compressing the information of multiple tokens into a lower dimension is not only an effective way to improve computational efficiency, but also an effective way to enhance the learning efficiency and representation ability of Transformers in some cases.**

## 5. Conclusions and Future Work

Considering that existing Transformer-based multivariate time series forecasting methods struggle to effectively extract and fuse cross-variable and cross-time dependencies when faced with causal time lags between variables, we propose Sensorformer, which first compresses global information and then extracts dependencies and features through pure cross-patch attention. Experimental results show that Sensorformer achieves state-of-the-art (SOTA) performance in many cases. In the future, we plan to further explore how to enhance the ability of Transformers to handle multivariate time series forecasting tasks with continuously changing variable dimensions.

**Acknowledgments**

This research work was supported by the following funds and organizations: National Natural Science Foundation of China (Grant No. 62073342). National Natural Science Foundation of China (Grant No. U23A20329). Fundamental Research Funds for the Central Universities of Central South University.(No. 2024ZZTS0099). High Performance Computing Center of Central South University.

## Appendix A. Experimental Details
### A.1 Datasets and Baselines

**Dataset:** We conducted experiments on 9 mainstream real-world datasets following iTransformer[13]. (1) ETTh1 and ETTh2 (Electricity Transformer temperature-hourly); (2) ETTm1 and ETTm2 (Electricity Transformer temperature-minutely); (3) Weather; (4) ECL (Electricity Consuming Load); (5) Exchange; (6) Traffic; (7) Solar-Energy. The detailed information of all datasets is shown in Table 3. The physical meanings of each dataset are as follows: (1) The ETT dataset (including ETTh1/2 and ETTm1/2) records the changes of 7 variables of transformers from 2016 to 2016; (2) The Exchange dataset records the daily exchange rate changes of 8 countries from 1990 to 2016; (3) The Weather dataset records the changes of 21 meteorological factors collected by the Weather Station of the Max Planck Biogeochemistry Institute in 2020; (4) The ECL dataset records the hourly electricity consumption of 321 households; (5) The Traffic dataset records the road occupancy measured by 862 highway sensors in the San Francisco Bay area freeways; (6) The Solar-Energy dataset records the power generation of 137 photovoltaic power stations.

Table 3. Detail information of all datasets

| Dataset | Dim | Prediction Length | Train/Vali/Test Size |
|---|---|---|---|
| ETTh1/2 | 7 | {96,192,336,720} | (8545, 2881, 2881) |
| ETTm1/2 | 7 | {96,192,336,720} | (34465, 11521, 11521) |
| Exchange | 8 | {96,192,336,720} | (5120, 665, 1422) |
| Weather | 21 | {96,192,336,720} | (36792, 5271, 10540) |
| ECL | 321 | {96,192,336,720} | (18317, 2633, 5261) |
| Traffic | 862 | {96,192,336,720} | (12185, 1757, 3509) |
| Solar-Energy | 137 | {96,192,336,720} | (36601, 5161, 10417) |

**Baselines:** We adopted the 9 mainstream baseline methods and reproduced the results in our own experimental environment based on the implementations in the Time-series-library[32] and the results reported in paper[13], including (1) Transformer-based methods: iTransformer[13], PatchTST[8], Autoformer[26], Crossformer[9], FEDformer[28], Stationary[35]. (2) Linear-based methods: TiDE[34], D-Linear[7]. (3) TCN-based methods: TimesNet[18], SCINet[19].

### A.2 Experimental environment and key hyperparameter details

All experimental code was implemented using PyTorch and run on an NVIDIA RTX 3090ti 24GB GPU. All baseline methods used the ADAM optimizer with an initial learning rate of 0.0001, the training epochs was fixed to 10, and batch size was fixed to 32. Loss function was MSE Loss. For all datasets, the lookback window was fixed to 96, the prediction length $H \in \{96,192,336,720\}$, and the training epoch was fixed to 10. The train/validation/test split ratios for each dataset follow the division method used in paper[13]. The length of representation $d\_model$ was fixed to 256, the number of Sensor Attention Blocks(encoder layers for other transformer-based methods) was fixed to 2, the number of attention heads was fixed to 2, the patch length was fixed to 32, and the patch stride was fixed to 8. We used the code from the Time-series-library[32] to regenerate the results of iTransformer, PatchTST and Crossformer, ensuring that the hyperparameters were consistent with those of Sensorformer. The Time-series-library is an objective third-party code library for multivariate time series forecasting methods. The training code and result evaluation code for Sensorformer also used the Time-series-library to ensure consistency in statistical standards and objectivity of results. For the results of other baseline methods besides iTransformer, PatchTST and Crossformer, we referred to the results reported in [13] for comparison.

**Appendix B. Other Experimental Results.**

**B.1 Complete experimental results of causal time lag analysis between variables**

To visually present the distribution of causal time lags between variables in different datasets, we conducted the following experiments:

**Exp1.** Proportion of variable pairs with causal time lag, the calculation process is described in Section 4.2. The experimental results are shown in Table 4 and Fig.6.

**Exp2.** Distribution of casual time lag distance in different datasets, with the specific calculation method as follows: (1) Randomly select an input tensor; (2) After performing patching operations on the input tensor, we take the middle patch of the i-th variable (the 5-th patch for all datasets due to consistent dataset division) in this tensor as the base patch, then calculate the PCC between the base patch and all patches of the j-th variable; (3) Calculate the absolute value of the difference between the base patch and the patch with maximum PCC of the j-th variable, this difference is referred to as the causal time lag distance; (4) calculate the average causal time lag distance between all variable pairs in the current input tensor; (5) repeat the (1)-(4) 10 times for each dataset. The experimental results are shown in Table 5 and Fig.7.

Table 4. Proportion of variable pairs with causal time lag across different datasets

| Datasets | ETTm1 | ETTh1 | Exchange | Weather | ECL | Traffic | Solar-Energy |
|---|---|---|---|---|---|---|---|
| tensor 1 | 66.67% | 80.9% | 50.00% | 54.29% | 72.34% | 75.71% | 48.00% |
| tensor 2 | 71.23% | 85.71% | 57.14% | 71.42% | 81.92% | 61.18% | 100% |
| tensor 3 | 47.62% | 80.22% | 78.13% | 73.33% | 63.88% | 16.10% | 81.45% |
| tensor 4 | 66.67% | 28.13% | 64.23% | 85.71% | 70.6% | 75.24% | 6.93% |
| tensor 5 | 52.36% | 66.67% | 64.28% | 63.34% | 71.05% | 52.32% | 33.17% |
| tensor 6 | 57.14% | 33.33% | 82.11% | 48.71% | 74.18% | 24.46% | 22.20% |
| tensor 7 | 60.32% | 66.67% | 32.71% | 71.65% | 68.36% | 70.55% | 27.68% |
| tensor 8 | 71.3% | 71.42% | 33.25% | 58.12% | 59.10% | 56.38% | 76.21% |
| tensor 9 | 66.67% | 29.24% | 46.45% | 77.6% | 75.88% | 40.21% | 55.4% |
| tensor 10 | 76.19% | 70.6% | 70.08% | 67.13% | 82.39% | 66.24% | 63.19% |
| Avg | 63.62% | 61.29% | 57.84% | 67.13% | 71.97% | 53.84% | 51.42% |

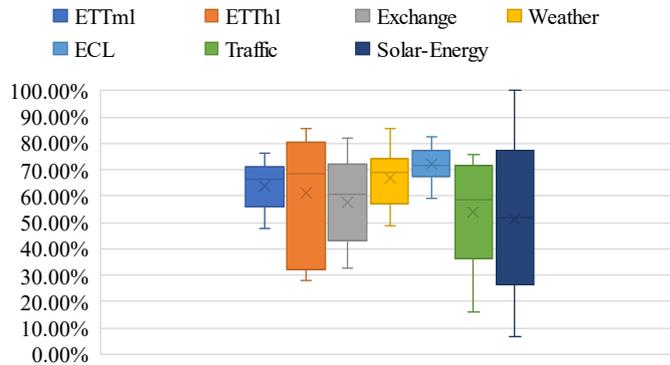

Fig. 6 Distribution of proportion of variable pairs with causal time lag

Table 5. Average causal time lag distance across different datasets

| Datasets | ETTm1 | ETTh1 | Exchange | Weather | ECL | Traffic | Solar-Energy |
|---|---|---|---|---|---|---|---|
| tensor 1 | 1.52 | 2.62 | 2.36 | 2.21 | 2.25 | 1.72 | 0.02 |
| tensor 2 | 1.23 | 1 | 2.32 | 1.77 | 2.36 | 1.62 | 5.00 |
| tensor 3 | 0.81 | 1.62 | 1.46 | 2.56 | 2.29 | 2.10 | 0.18 |
| tensor 4 | 2.48 | 2.42 | 1.79 | 1.76 | 2.3 | 2.00 | 0.48 |
| tensor 5 | 1.9 | 1.48 | 1.5 | 2.96 | 2.4 | 2.15 | 5.00 |
| tensor 6 | 2.23 | 1.53 | 2.14 | 1.92 | 2.17 | 1.98 | 0.25 |
| tensor 7 | 2.38 | 2.05 | 2 | 2.65 | 2.3 | 1.64 | 2.71 |
| tensor 8 | 2.37 | 2.24 | 1.46 | 2.23 | 2.34 | 1.69 | 0.40 |
| tensor 9 | 2.47 | 2.42 | 1.29 | 3.03 | 2.21 | 2.20 | 0.10 |
| tensor 10 | 2.33 | 1.3 | 1.86 | 2.88 | 2.42 | 2.24 | 1.40 |
| Avg | 1.97 | 1.87 | 1.82 | 2.40 | 2.30 | 1.93 | 1.55 |

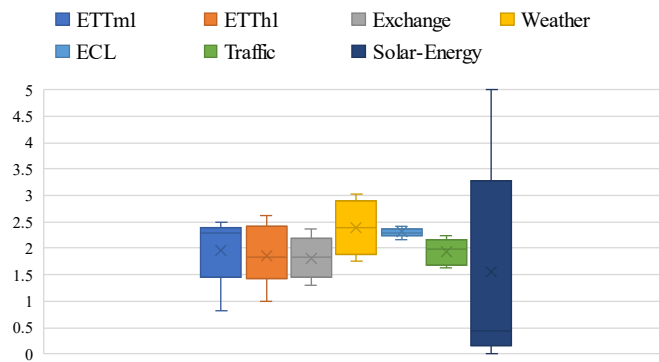

Fig. 7 Distribution of average causal time lag distance

Observing the experimental results, we can draw the following main findings:

(1) The distribution of causal time lags varies significantly between different datasets. Fig. 6 shows that in the Solar-Energy and ETTm1 datasets, the dynamic causal time lag distances within different input tensors vary greatly (with a large standard deviation in the boxplot). This indicates that the causal time lag between variables may exhibit strong dynamics in many cases, revealing the limitations of static evaluation and alignment of time lags between variables during the preprocessing stage. Incorporating dynamic estimation of time lags between variables during network inference is an important technique. This is precisely the capability that Sensorformer, which employs pure cross-patch attention, inherently possesses.

(2) The ETTm1 and ETTh1 datasets have identical physical meanings, but due to different sampling frequencies, the distribution of time lags between variables shows significant differences. We believe the main reason for this phenomenon is that when the sampling frequency is greatly changed, the scale of causal relationships observed under the same receptive field will change significantly. At high sampling frequencies, the causal relationships or correlations observed under the same receptive field are dominated by short-term or local factors, while at low sampling frequencies, they are dominated by long-term or global factors. Therefore, when analyzing multivariate time series with different sampling frequencies, it is necessary to consider the impact of short-term and long-term factors due to sampling frequency.

**B.2 Effect of Hyperparameters**

Since Sensorformer is a patch-based multivariate time series forecasting method, the way patches are divided is a major factor affecting its representation and learning capabilities. Therefore, this section mainly discusses the impact of three hyperparameters on Sensorformer's performance: Patch length,

Patch stride (i.e., the degree of overlap between patches), and $d\_model$ (the length of representation vector of a patch). The experimental parameters are fixed with a lookback window of 96 and a prediction length of 96, with other training parameters consistent with Appendix A.2. The following adjustments and comparisons are then made: (1) In the Patch length impact experiment, we fix Patch stride=8 and $d\_model$=256, and compare the differences in MSE when Patch length is set to $P \in \{8,16,32,64\}$ on the ETTh1 (low-dimensional dataset), Weather (medium-dimensional dataset) and ECL (high-dimensional dataset) datasets. (2) In the Patch stride impact experiment, we fix Patch length=32 and $d\_model$=256, and compare the differences in MSE when Patch stride is set to $S \in \{8,16,32,64\}$. (3) In the $d\_model$ impact experiment, we fix Patch length=32 and Patch stride=8, and compare the differences in MSE when d_model is set to $d\_model \in \{64,128,256,512\}$. The experimental results are shown in Fig. 8.

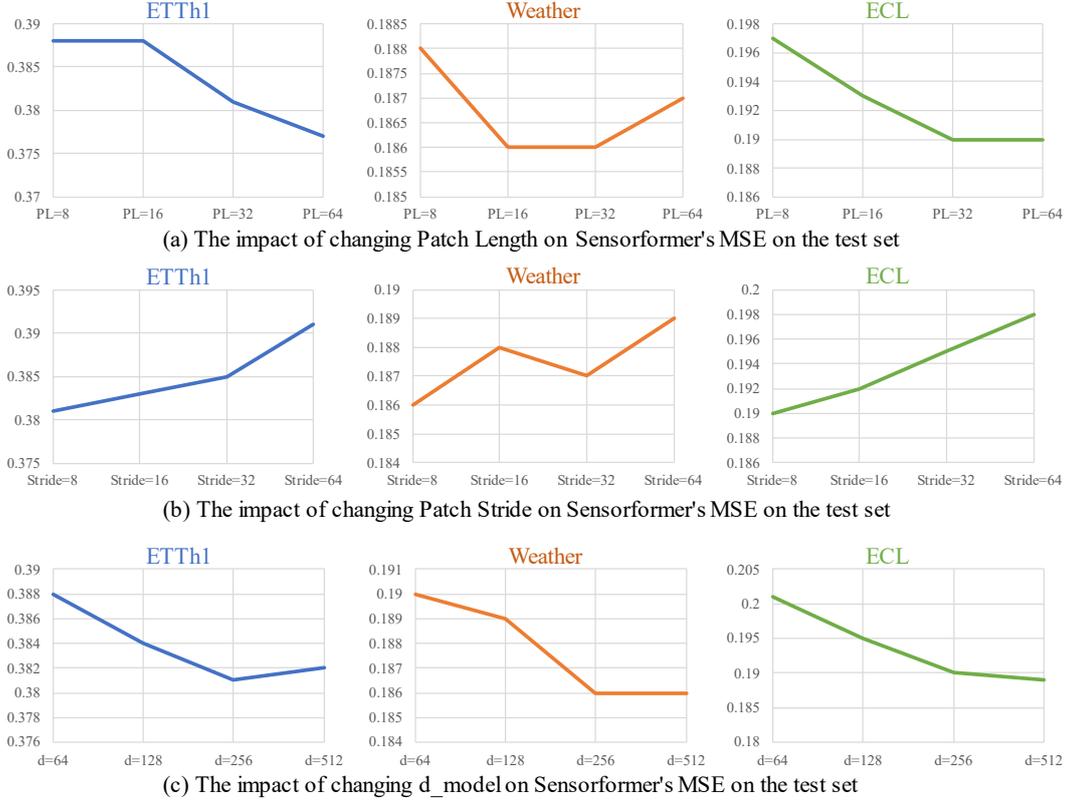

Fig. 8. Experimental results on the impact of changing hyperparameters on Sensorformer's performance

From Fig. 8, we can observe the following phenomena: (1) As the Patch length increases from 8 to 32, the MSE of Sensorformer on all datasets shows a significant decrease. However, when the Patch length increases from 32 to 64, the MSE no longer shows a significant decrease. This indicates that neither too small nor too large Patch length is suitable for Sensorformer, and a reasonable Patch length should be around one-third of the lookback window. (2) When the Patch stride is 8 and 16, meaning there is some overlap between patches, the MSE of Sensorformer on both datasets is relatively low. However, when the Patch stride is 32, meaning there is no overlap between patches, the MSE on ETTh1 and ECL increases significantly. When the Patch stride is 64, meaning some data points are discarded after patching, the MSE shows a significant increase on all datasets. This indicates that when setting the Patch stride, there should be a certain degree of overlap between patches. This is an interesting issue worth further theoretical exploration, as overlap seems to introduce more redundant data, increasing the computation complexity. However, in practice, a certain degree of data redundancy can lead to better performance. (3) As $d\_model$ increases, the prediction error of Sensorformer on all datasets continues to decrease. However, due to limited computational resources, we were unable to observe at what point the performance starts to decline as $d\_model$ increases. Nevertheless, the decrease in MSE from $d\_model$ increasing from 256 to 512 is less than the decrease from 128 to 256. Since increasing $d\_model$ linearly increases the computational complexity of the network, in practical applications, when Patch length is 32, setting $d\_model$ to around 256 may be a good trade-off between accuracy and computational efficiency.

**B.3 Performance Robustness Evaluation of Sensorformer.**

To verify the stable superiority of Sensorformer's architecture, we randomly selected 5 random seeds and tested the final learning performance of Sensorformer under 5 different random initializations. The experimental results are shown in Table 6. It can be seen that Sensorformer's final learning performance is almost identical under different random initialization parameters (with very small standard deviations), indicating that Sensorformer's performance superiority is robust.

**Table 6. Performance robustness of Sensorformer.**

| Dataset | ETTh1 | | ECL | | Exchange | | Weather | |
|---|---|---|---|---|---|---|---|---|
| Horizon | MSE | MAE | MSE | MAE | MSE | MAE | MSE | MAE |
| 96 | 0.381±0.000 | 0.4±0.001 | 0.19±0.001 | 0.279±0.000 | 0.086±0.001 | 0.204±0.001 | 0.186±0.000 | 0.223±0.001 |
| 192 | 0.43±0.001 | 0.428±0.001 | 0.195±0.001 | 0.285±0.001 | 0.184±0.001 | 0.302±0.001 | 0.232±0.001 | 0.263±0.001 |
| 336 | 0.465±0.002 | 0.446±0.001 | 0.212±0.002 | 0.302±0.001 | 0.334±0.001 | 0.418±0.001 | 0.285±0.002 | 0.302±0.001 |
| 720 | 0.504±0.004 | 0.488±0.002 | 0.254±0.005 | 0.334±0.006 | 0.851±0.010 | 0.697±0.006 | 0.358±0.002 | 0.35±0.001 |

**B.4 Visualization of prediction results**

Figures 9 to 15 compare the differences in trend between the predicted value curves and the true value curves of Sensorformer, iTransformer, and PatchTST on six datasets.

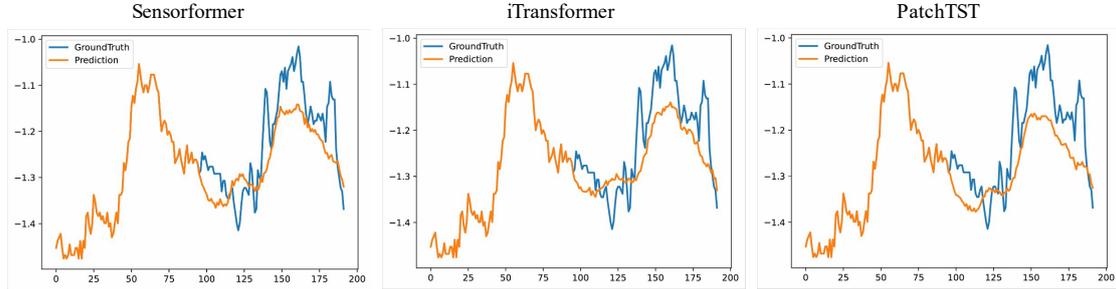

Fig. 9 Prediction Results on the ETTm1 Dataset with Input-96 and Predict-96.

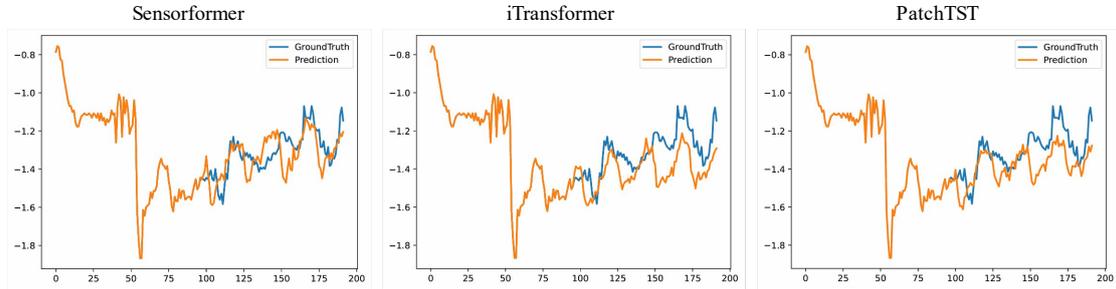

Fig. 10 Prediction Results on the ETTh1 Dataset with Input-96 and Predict-96.

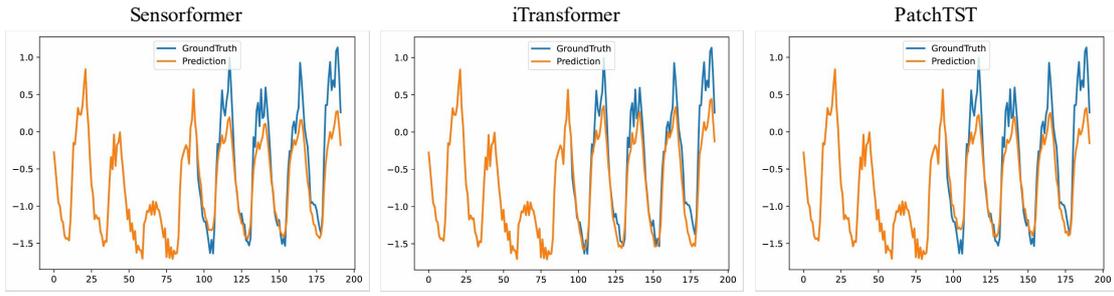

Fig. 11 Prediction Results on the ECL Dataset with Input-96 and Predict-96.

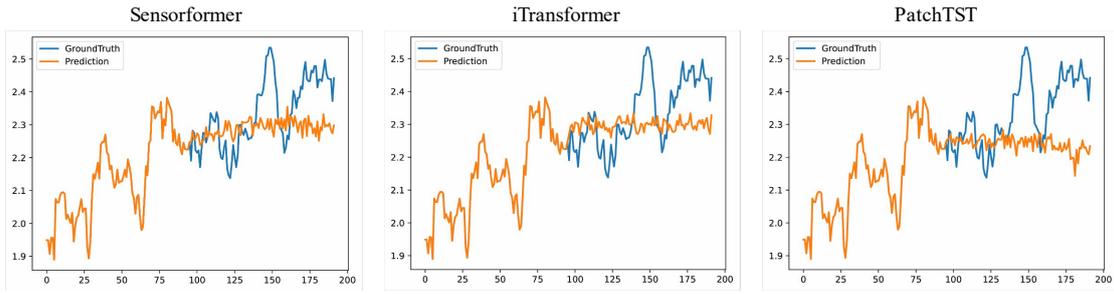

Fig. 12 Prediction Results on the Exchange Dataset with Input-96 and Predict-96.

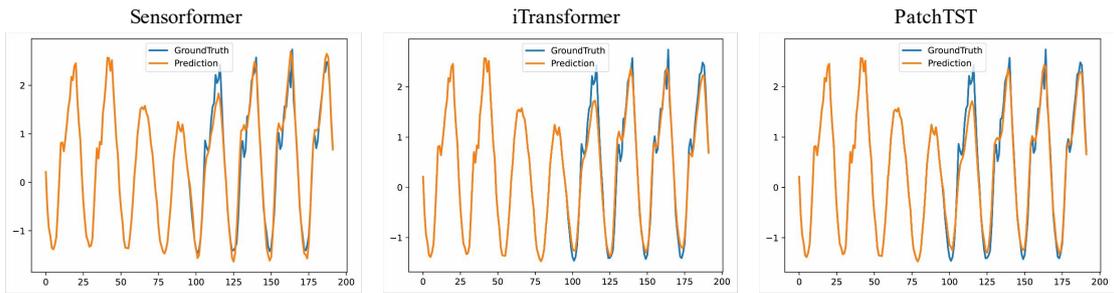

Fig. 13 Prediction Results on the Traffic Dataset with Input-96 and Predict-96.

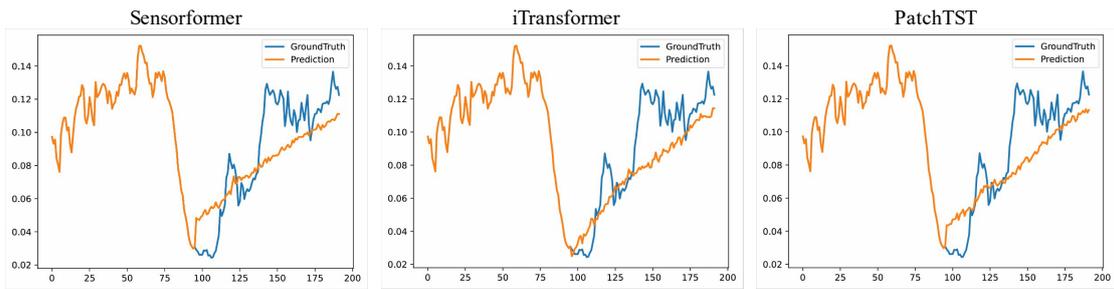

Fig. 14 Prediction Results on the Weather Dataset with Input-96 and Predict-96.

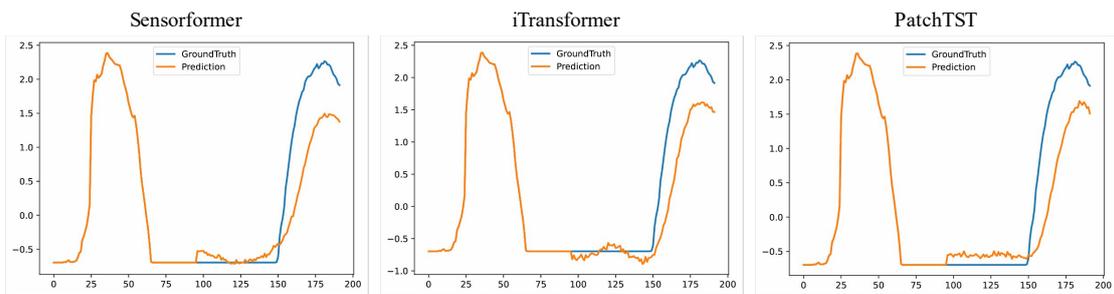

Fig. 15 Prediction Results on the Solar-Energy Dataset with Input-96 and Predict-96.